\pdfoutput=1

\documentclass[11pt]{article}

\usepackage[]{acl}

\usepackage{times}
\usepackage{latexsym}
\usepackage{multirow}
\usepackage{color,xcolor}
\usepackage[T1]{fontenc}
\usepackage{booktabs} 
\usepackage[utf8]{inputenc}

\usepackage{microtype}

\usepackage{inconsolata}
\usepackage{amsmath}
\usepackage{subcaption}
\usepackage{graphicx}

\title{CitaLaw: Enhancing LLM with Citations in Legal Domain}

\author{
    Kepu Zhang\textsuperscript{1},
    Weijie Yu\textsuperscript{2}\thanks{~~Corresponding author},
    Sunhao Dai\textsuperscript{1},
    Jun Xu\textsuperscript{1} \\
    \textsuperscript{1}Gaoling School of Artificial Intelligence, Renmin University of China \\
    \textsuperscript{2} University of International Business and Economics \\
    \texttt{kepuzhang@ruc.edu.cn } 
    \\
}

\begin{document}
\maketitle
\begin{abstract}
In this paper,
we propose \texttt{CitaLaw}, the first benchmark designed to evaluate LLMs’ ability to produce legally sound responses with appropriate citations. \texttt{CitaLaw} features a diverse set of legal questions for both laypersons and practitioners, paired with a comprehensive corpus of law articles and precedent cases as a reference pool. 
This framework enables LLM-based systems to retrieve supporting citations from the reference corpus and align these citations with the corresponding sentences in their responses.
Moreover, we introduce syllogism-inspired evaluation methods to assess the legal alignment between retrieved references and LLM-generated responses, as well as their consistency with user questions. 
 Extensive experiments on 2 open-domain and 7 legal-specific LLMs demonstrate that integrating legal references substantially enhances response quality. Furthermore, our proposed syllogism-based evaluation method exhibits strong agreement with human judgments.
\end{abstract}

\section{Introduction}
Generating responses supported by citations, such as relevant law articles and precedent cases, is essential for ensuring the trustworthiness of large language models (LLMs) in legal tasks. For laypersons seeking legal advice~\cite{fei2023lawbench}, LLM-generated responses grounded in citations provide verifiable information, fostering trust in the system. Conversely, for legal practitioners such as lawyers and judges, citations serve as supportive evidence that aids in analyzing complex cases, validating legal arguments, and ensuring decisions align with established legal principles~\cite{li2024lexevalcomprehensivechineselegal,zhong2020jec,abdallah2023exploring}. 

Recently, a growing body of benchmark research~\cite{gao2023rarr,li2023survey} has focused on enabling LLMs to provide citations for the statements they generate.
For instance, 
ALCE~\cite{gao2023enabling} introduces a benchmark designed to evaluate the ability of LLMs to generate citation-supported outputs, aiming to improve factual accuracy. WebCiteS~\cite{deng2024webcites} provides a curated database of manually annotated summaries and citations to enhance performance in text summarization and citation generation. 

\begin{figure}
    \centering
\includegraphics[width=0.98\linewidth]{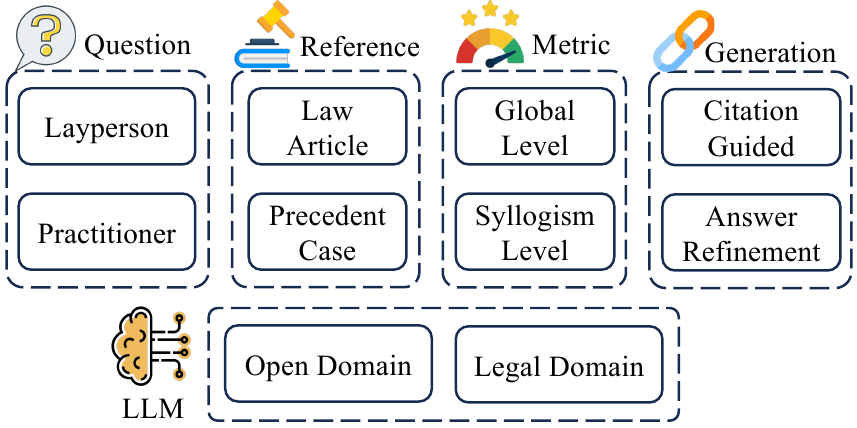}
    \caption{
    The framework of our \texttt{CitaLaw}.
            }
    \label{fig:source}
    \vspace{-3mm}
\end{figure}

\begin{figure*}
    \centering
\includegraphics[width=0.98\linewidth]{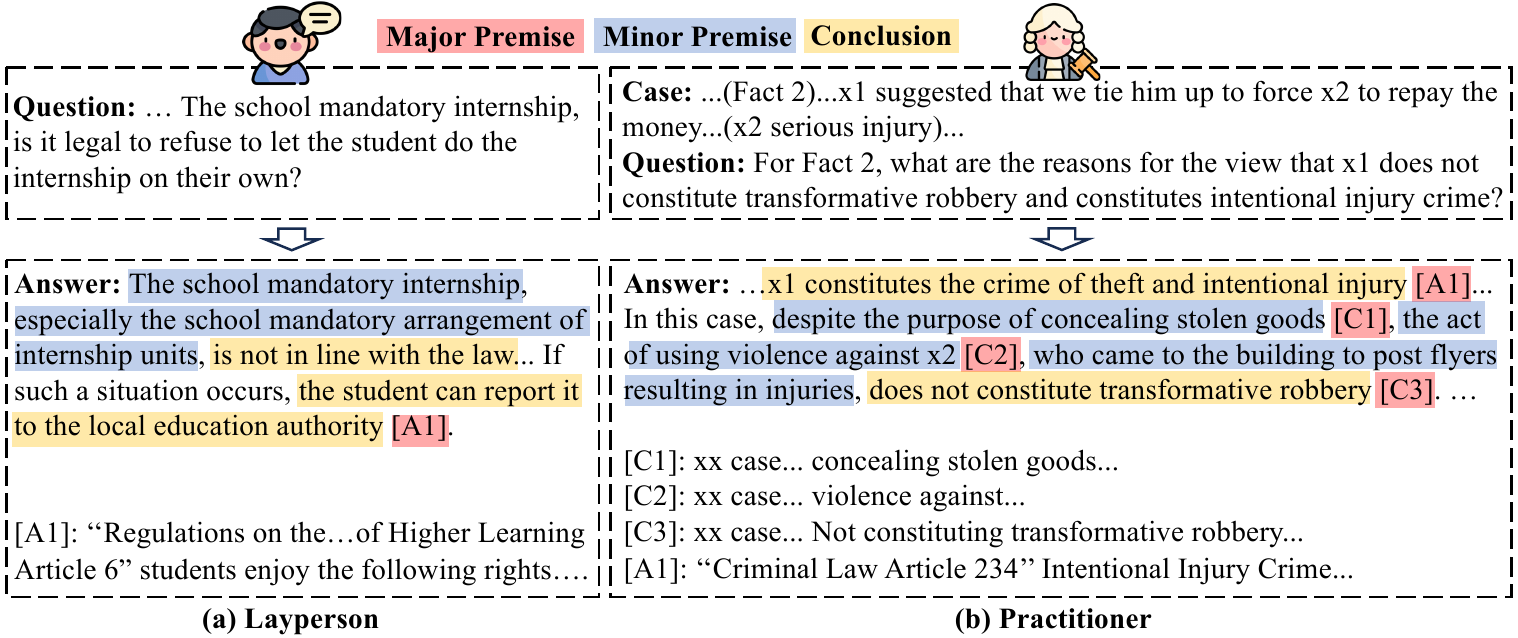}
    \caption{
    Examples from the two subsets of \texttt{CitaLaw}, with text in red, blue, and yellow representing the three dimensions of the syllogism: major premise, minor premise (circumstances, illegal acts), and conclusion (legal decisions), respectively. [A] and [C] denote citations to relevant law articles and precedent cases, respectively.
            }
    \label{fig:example}
    \vspace{-3mm}
\end{figure*}

While these studies have made notable progress in general domains, they face significant challenges when applied to the legal domain.
\textbf{First}, 
laypersons and legal practitioners interact with LLMs differently and have distinct expectations for citations. Laypersons typically seek legal advice and rely on citations to verify the accuracy of LLM responses, whereas legal practitioners pose more complex queries, using LLMs for legal reasoning, with citations serving as supportive evidence. Existing studies fail to address these differences, leading to unsatisfactory performance in real-world applications.
\textbf{Second}, 
existing methods often fall short in providing the diverse references required in legal contexts, such as law articles and precedent cases. Law articles establish the foundational legal framework, while precedent cases offer concrete examples and interpretive guidance. These two types of references inherently align with the distinct characteristics of civil and common law systems.
\textbf{Third}, traditional citation evaluation measures, such as ROUGE~\cite{lin2004rouge}, rely on surface-level similarities and are often insufficient to assess the alignment between references and LLM-generated responses.
In the legal domain, effective evaluation requires a deeper understanding of logical and semantic relationships.

To overcome the above challenges, we propose \texttt{CitaLaw}, the \textbf{first} benchmark tailored to evaluate LLMs' capabilities in generating legally grounded responses supported by accurate and context-aware citations. As shown in~\autoref{fig:source}, \texttt{CitaLaw} incorporates four distinct legal-specific features:

(1) \texttt{CitaLaw} has two subsets tailored for laypersons and practitioners, with examples in Figure~\ref{fig:example}.
Laypersons typically ask shorter, conversational questions, while practitioners often pose specialized, detailed questions.

(2) \texttt{CitaLaw} includes a retrieval corpus comprising two commonly used references: law articles, which provide clear and concise guidelines for addressing user questions, and precedent cases, which offer legal reasoning and support for judicial decisions. 
Recognizing the distinct needs of laypersons and practitioners, we provide only law articles for laypersons to ensure clarity, while practitioners have access to both law articles and precedent cases to support more complex legal reasoning.

(3) In addition to traditional global-level metrics such as MAUVE~\cite{pillutla2021mauve}, we propose a syllogism-based evaluation method to assess both the response correctness and the citation quality. This method provides a more granular evaluation by focusing on three key dimensions: circumstances, illegal acts, and legal decisions.

(4) We consider two types of response generation methods. The first type, Citation-Guided Generation (CGG), involves generating responses by incorporating retrieved references during generation. The second type, Answer Refinement Generation (ARG), refines the LLMs’ initial response (CloseBook) by retrieving and incorporating reference information. This category includes ARG-Q, which retrieves citations using only the user query, and ARG-QA, which retrieves citations using both the user query and the LLM’s initial response.

Extensive experiments on two open-domain and seven legal-specific LLMs reveal the following key insights:
1) Incorporating legal references into the LLM significantly improves the quality of responses;
2) Including references as part of the LLM’s input consistently outperforms answer-refinement methods;
3) Leveraging references to refine the LLM’s responses yields better alignment of responses and
references.
4) For fine-tuning LLMs in legal scenarios, incorporating law articles, syllogistic reasoning, and full-scale fine-tuning achieves promising performance.
5) Open-domain LLMs surprisingly outperform legal-specific LLMs in certain scenarios;
6) Human evaluations show a strong correlation with our syllogism-based methods.

In summary, our contributions are as follows:
\begin{itemize}
    \item 
    To the best of our knowledge,  \texttt{CitaLaw} is the \textbf{first} benchmark designed to evaluate the capability of LLMs to generate legally grounded responses with accurate and context-aware citations. \texttt{CitaLaw} includes questions tailored to both laypersons and practitioners, paired with a citation corpus comprising law articles and precedent cases.
    \item We propose a two-level evaluation framework that combines global-level metrics with a syllogism-based reasoning approach. 
    Additionally, we explore two mainstream methods for legal response generation: citation-guided and answer refinement.
    \item 
    Through extensive experiments on two open-domain and seven legal-specific LLMs, we demonstrate the effectiveness of integrating legal references into response generation and validate our syllogism-based evaluation method. Additionally, we provide actionable insights for the practical deployment of LLMs in legal scenarios.
\end{itemize}

\section{Related Work}

\textbf{LLM for Legal Task.}
A amount of work has explored applying LLMs to legal tasks~\cite{savelka2023explaining,wu2023precedent,yu2022legal,blair2023can}. 
Building LLMs tailored for legal scenarios is a popular direction~\cite{yue2023disclawllm,sdu_fuzi_mingcha,HanFei}. 
There are also some benchmarks that explore the capabilities of LLMs in legal tasks. LawBench~\cite{fei2023lawbench} evaluates LLMs' legal knowledge across three cognitive aspects. LAiW~\cite{dai2023laiw} assesses LLMs' legal reasoning abilities based on legal practice logic. LexEval~\cite{li2024lexevalcomprehensivechineselegal} evaluates LLMs' legal capabilities based on a new legal cognitive ability classification system.
However, none of them have considered enhancing the trustworthiness of LLMs in legal scenarios by generating outputs with citations.

\textbf{Citation in LLM.}
Attribution~\cite{li2023survey} in LLMs refers to providing supporting evidence for the answers generated by the model, presented in the form of citations. ALCE~\cite{gao2023enabling} is an automated benchmark for evaluating LLMs' ability to generate outputs with citations, aimed at improving the factual accuracy of the generated responses. WebCiteS~\cite{deng2024webcites} provides a database containing 7,000 manually annotated summaries and citations to enhance LLMs' capabilities in summarization and citation. RARR~\cite{gao2023rarr} enhances LLM outputs by automatically adding citations, and modifying the responses. ExpertQA~\cite{malaviya2024expertqa} verifies and modifies citations through expert review to ensure reliability.
In contrast to the above works, \texttt{CitaLaw} focuses specifically on citation in legal scenarios.

\section{Task Setup and Dataset Construction}\label{sec:task and dataset}
Suppose we have a legal corpus \( D \), which consists of either a collection of precedent cases $(D_l)$ or law articles $(D_c)$. Given a user question \( x \) posed by either a layperson or a practitioner, the LLM-based system is tasked with retrieving supportive citations from \( D \) and generating a legally grounded response \( y \). The response \( y \) comprises a list of \( n \) sentences, i.e., \( y = [s_1, \cdots, s_n] \), where each sentence \( s_i \) refers to at most one corresponding citation. As illustrated in~\autoref{fig:example}, the system is further required to attach each citation to its relevant sentence, with “[A]” and “[C]” denoting references to law articles and precedent cases, respectively.

To enable the evaluation of this task, we construct the specialized dataset (Table~\ref{tab:data_main} shows the statistics) as follows:

To simulate the behavior of \textbf{laypersons}, we include questions that are more conversational, lack detailed case descriptions, and are relatively short in length. We use the consultation section from LawBench~\cite{fei2023lawbench}, which collects user queries from the Hualv website\footnote{www.66law.com} and answers provided by lawyers or legal consulting firms.

To simulate the behavior of \textbf{legal practitioners}, we include questions that are more professional, often accompanied by detailed case descriptions, and generally longer. For this purpose, we use the open-ended question section from LexEval~\cite{li2024lexevalcomprehensivechineselegal}, which consists of subjective questions from the National Uniform Legal Profession Qualification Examination. These questions are particularly challenging for LLMs, requiring them to understand the case fully and apply legal knowledge accurately to generate answers.

In terms of the \textbf{corpus}, we construct a comprehensive corpus from multiple sources, including law articles and precedent cases. Specifically, for law articles, we collect approximately 50,000 documents from LexiLaw\footnote{https://github.com/CSHaitao/LexiLaw}, covering areas such as Civil Law, Criminal Law, and judicial interpretations. For precedent cases, we include both criminal and civil cases. Criminal cases are sourced from the LeCaRD legal retrieval dataset~\cite{ma2021lecard}, ELAM~\cite{yu2022explainable}, and civil cases from the CAIL legal summary dataset, LJP-MSJudge~\cite{ma2021legal}, and the pre-training data of fuzi.mingcha~\cite{sdu_fuzi_mingcha}. As a supplement to precedent cases, we also incorporate question-and-answer pairs from fine-tuning datasets of legal LLMs as part of the precedent cases. These QA pairs are collected from DISC-LawLLM~\cite{yue2023disclawllm}, LawGPT\_zh~\cite{LAWGPT-zh}, and HanFei~\cite{HanFei}. In total, the constructed corpus contains approximately 500,000 documents, ensuring sufficient coverage of both law articles and precedent cases to support diverse legal tasks.

\begin{table}[t]
\centering
\resizebox{0.5\textwidth}{!}{
\begin{tabular}{l|cccc}
    \toprule
    Dataset &\#Q & Len$_Q$ & Len$_A$ & Q Type  \\
    \hline
    Layperson &500 &57.62&107.40&Question\\ 
    \hline
    Practitioner&500 &618.96&193.46 &Case + Question\\
    \bottomrule
\end{tabular}
}
\caption{
Dataset statistics. \#Q indicates the number of questions, Len$_Q$ and Len$_A$ denote the average lengths of questions and gold answers, and Q Type refers to the question type.
}
\label{tab:data_main}
\end{table}

\section{Method}\label{sec:method}
\subsection{Response Generation}
We consider two types of methods in this study.

\textbf{Citation-Guided Generation (CGG)} 
produces response $y_{cgg}$ given a user question $x$ by referring retrieved relevant document(s) $D_R$:
\begin{equation}
\label{eq:type1}
   y_\mathrm{cgg}=f_{\mathrm{LLM}}(x,D_R,p_1),
\end{equation}
where $f_{\mathrm{LLM}}$ denotes a open-domain or a legal specific LLM; $p_1$ is the direct generation prompt. All prompt settings are detailed in~\autoref{sec:appendix:prompt}. 

\textbf{Answer Refinement Generation (ARG)} 
is a two-stage method that generates the final response $y_\mathrm{arg}$ by refining the LLM’s initial response $y_\mathrm{init}$ through the retrieval and incorporation of reference information. This process can be formulated as:
\begin{equation}
\label{eq:type2}
y_\mathrm{init}=f_{\mathrm{LLM}}(x,p_2),
\end{equation}
where $p_2$ is the prompt instructing the LLM to directly generate an initial response without reference information. We refer to this step as \textbf{CloseBook}. The initial response  $y_\mathrm{init}$ is then refined as:
\begin{equation}
\label{eq:type3}
    y_\mathrm{arg}=f_{\mathrm{LLM}}(y_\mathrm{init},D_R,p_3),
\end{equation}
where $p_3$ is the prompt guiding the LLM to refine the $y_\mathrm{init}$ using the retrieved documents $D_R$.

Laypersons and practitioners interact with LLMs differently and have distinct expectations for citations. When  $x$  is submitted by a layperson, the corresponding  $D_R$  consists of relevant law articles. In contrast, when $x$ is submitted by a practitioner, the corresponding  $D_R$  includes both relevant law articles and precedent cases. The process for retrieving  $D_R$  from $D$  is detailed in the next subsection.
\subsection{Citation Retrieval}
We explore state-of-the-art open-domain dense retriever BGE~\cite{xiao2023c}, along with two legal-specific dense retrievers, Criminal-BERT~\cite{zhong2019openclap} and Civil-BERT~\cite{zhong2019openclap}.
We also investigate two types of retrieval queries: $x$ (the user question alone, \textbf{ARG-Q}) and $[x;y_\mathrm{init}]$ (the concatenation of the user query  $x$  and the initial response  $y_\mathrm{init}$, where  [;]  denotes the concatenation operation, \textbf{ARG-QA}). The impact of different retrieval models on performance will be analyzed in the experiments.

\subsection{Citation Attachment}
Building on the retrieved citations, this subsection outlines the process of attaching these law articles or precedents to specific sentences in the LLM-generated responses. This process involves answering two key questions:

\textbf{What kind of sentences can be associated with citations?} 
We utilize co-occurring words and legal entity extraction to identify sentences that explicitly reference legal concepts, actions, or terms relevant to the retrieved citations. Specifically, we construct a pool of legal terminologies using THUOCL\footnote{https://github.com/thunlp/THUOCL} and LaWGPT~\cite{lawgpt}. 
A sentence is considered eligible if it contains any of the terminologies from this pool. Additionally, we use SpaCy~\cite{Honnibal_spaCy_Industrial-strength_Natural_2020} to extract legal entities from each sentence. If a sentence includes legal entities, it is also deemed eligible for citation attachment.

\textbf{How are citations attached to the identified sentences?}
If a sentence is deemed eligible for citation attachment, we associate it with retrieved citations as follows.
For the laypersons, the retrieved law article $c_l\in D_l$ is attached to the most relevant sentence $s_k\in y$: 
\begin{equation}
\label{eq:c_common}
C_{\mathrm{Lay}}=\{(s_k, c_l) \mid s_k = \mathop{\arg\max}_{s_i \in y} \mathrm{sim}(s_i, c_l)\},
\end{equation}
where $(s_k, c_l)$ represents attaching the reference $c_l$ to the sentence $s_i$, and $\mathrm{sim}(\cdot)$ is computed using sentence-BERT~\cite{reimers2019sentence}. We set $|C_{\mathrm{Lay}}|=1$ because, typically, a layperson’s query pertains to only one specific legal article.
For practitioners, we attach the retrieved law article in the same way as for laypersons. Additionally, we associate the retrieved precedent cases $c_c\in D_c$ with each $s_i\in y$, which is formulated as:
\begin{equation}
\begin{aligned}
\label{eq:c_legal}
C_{\mathrm{Pra}}=&\{(s_k, c_l) \mid s_k = \mathop{\arg\max}_{s_i \in y} \mathrm{sim}(s_i, c_l)\}
\\\cup&\{(s_i,c_c)|, c_c=\mathop{\arg\max}_{c_j\in D_c} \mathrm{sim}(s_i,c_j)\},
\end{aligned}
\end{equation}
where $|D_c|=3$, meaning each response $y$ can be associated with up to three precedents\footnote{Considering the input window size of LLMs, we set up to retrieve 3 precedent cases.}.

\section{Evaluation}\label{sec:eval}

\texttt{CitaLaw} provides a comprehensive evaluation framework incorporating metrics for fluency, correctness, and citation quality. This framework is divided into two levels of analysis: global level and the proposed syllogism level.

Syllogism, a foundational framework in legal reasoning, comprises three key components: the major premise, the minor premise, and the conclusion. In our legal context, these correspond to the relevant law article or precedent case (major premise), the factual circumstances and actions of a specific case (minor premise), and the resulting legal decision (conclusion). By integrating this syllogistic framework, \texttt{CitaLaw} goes beyond surface-level correctness to evaluate the logical coherence and alignment of LLM-generated responses with established legal principles.

\subsection{Fluency (Style Consistency)}
To ensure the LLM-generated responses align with the user’s requirements, the system must adapt its style based on the user’s background. For laypersons, responses should avoid excessive technical jargon to ensure accessibility and comprehension. Conversely, responses for legal practitioners should adopt a formal and professional tone to maintain credibility and utility.
To achieve this aim, we concatenate the user query and the LLM-generated response and apply MAUVE~\cite{pillutla2021mauve} to assess their style consistency.

\subsection{Correctness}
At the \textbf{global level}, we use established metrics ROUGE~\cite{lin2004rouge} and BERTScore~\cite{zhang2019bertscore}.  ROUGE measures word-level overlap between the generated and labeled responses, with scores reported for ROUGE-1, ROUGE-2, and ROUGE-L. BERTScore captures semantic similarity between the generated and labeled responses, and we report the F-score (BERT-F) for evaluation. These metrics assess the overall correctness of LLM-generated responses.

At the \textbf{syllogism level}, we leverage the Qwen2~\cite{qwen2} to extract key components, including the circumstances, illegal acts, and legal decisions. 
We use sentence-BERT~\cite{reimers2019sentence} to measure the alignment between the labeled responses and the generated outputs across these dimensions, resulting in Correct$_\mathrm{c}$, Correct$_\mathrm{a}$, and Correct$_\mathrm{d}$.
This syllogism-level evaluation allows us to assess the logical coherence of the responses, ensuring that they align with the underlying legal reasoning principles.

\subsection{Citation Quality}
As previously discussed, we assume that a question submitted by laypersons typically corresponds to a specific law article. Therefore, at the \textbf{global level}, we evaluate the citation quality of the retrieved law article (premise) by measuring its entailment with the associated sentence in the LLM’s response (hypothesis). Specifically, we use an NLI model to compute Cita$_\mathrm{Law}$, which quantifies the degree to which the law article entails the attached sentence. 
This metric reflects how effectively the response aligns with the cited law article. 
We employ DISC-LawLLM~\cite{yue2023disclawllm} as the NLI model due to its strong agreement with human evaluations (as discussed in Sec.~\ref{sec:human}) and its superior performance compared to other NLI models (as detailed in Sec.~\ref{sec:different nli}).

At the \textbf{syllogism level}, we evaluate the quality of precedent case citations by examining three key components: circumstances, illegal acts, and legal decisions. After extracting these elements from both the retrieved cases and the associated sentence in the LLM’s response, we utilize DISC-LawLLM to assess the entailment for each component. This evaluation yields three distinct scores: Cita$_\mathrm{c}$ for circumstances, Cita$_\mathrm{a}$ for illegal acts, and Cita$_\mathrm{d}$ for legal decisions, providing a more detailed and nuanced assessment of citation quality within the syllogism framework.

\section{Experiments}\label{sec:experiment}
\begin{table*}[h!]
    \resizebox{1\textwidth}{!}{
         \begin{tabular}{cl|c|ccc cccc|c|c}
          \toprule
          \multicolumn{2}{c|}{Metric} & \multicolumn{1}{c|}{Fluency } & \multicolumn{7}{c|}{Correctness} & \multicolumn{1}{c|}{Citation}  & \multicolumn{1}{c}{All}\\
          \hline
          Category & Model & Mauve & Rouge-1 & Rouge-2 & Rouge-L & BERT-F & Correct$_\mathrm{c}$ & Correct$_\mathrm{a}$ & Correct$_\mathrm{d}$  &  Cita$_\mathrm{Law}$ &\textbf{Avg }\\
          \hline
\multicolumn{1}{c}{\multirow {4}{*}{\shortstack{Llama3\\ (Llam3-8B-Instruct)}}}  
            &CloseBook & 22.63 & 16.47 & 1.95 & 13.34 & 58.46 & \textbf{73.05} & 68.24 & 66.87 & 67.38 &43.15 \\
&CGG & 61.01 & \textbf{23.97} & 6.05 & 17.91 & \textbf{65.94} & 67.29 & \textbf{77.31} & \textbf{74.95} & \textbf{86.70} &\textbf{53.46}\\
&ARG-Q & \textbf{61.27} & 23.17 & 5.65 & 17.83 & 64.23 & 69.04 & 75.45 & 74.47 & 79.10 & 52.24\\
&ARG-QA & 51.83 & 23.73 & \textbf{6.96} & \textbf{18.53} & 64.84 & 71.37 & 74.81 & 74.66 & 80.80 & 51.95 \\
\hline
          \multicolumn{1}{c}{\multirow {4}{*}{\shortstack{Qwen2\\(Qwen2-7B-Instruct)}}}  
          &CloseBook & 21.04 & 15.29 & 2.27 & 11.31 & 58.39 & \textbf{70.89} & 71.71 & 69.93 & 72.35 &43.69\\
&CGG & \textbf{75.10} & \textbf{22.26} & 4.77 & 15.41 & \textbf{65.28} & 67.50 & \textbf{78.62} & \textbf{77.82} & 77.59 &\textbf{53.82}\\
&ARG-Q & 66.55 & 20.86 & 4.50 & 15.42 & 64.59 & 66.96 & 77.82 & 75.66 & 81.48 & 52.65\\
&ARG-QA & 66.80 & 21.73 & \textbf{4.78} & \textbf{16.34} & 64.85 & 69.31 & 76.35 & 75.05 & \textbf{82.83} & 53.11\\
          \hline
          \multicolumn{1}{c}{\multirow {7}{*}{\shortstack{Legal LLM\\(CGG)}}}  
          &DISC-LawLLM & \textbf{72.70} & 22.46 & 4.14 & 15.48 & 65.06 & 65.21 & 78.55 & 76.17 & 83.46 &53.69\\
&fuzi.mingcha & 56.58 & 24.54 & 5.70 & 17.48 & \textbf{65.86} & 63.28 & \textbf{79.56} & \textbf{77.94} & 81.64 &52.51 \\
&LexiLaw & 71.89 & \textbf{24.96} & \textbf{6.25} & \textbf{18.91} & 65.68 & 68.89 & 78.12 & 76.72 & 82.42 &\textbf{54.87}\\
&Tailing & 13.95 & 15.93 & 4.13 & 12.89 & 59.47 & \textbf{72.00} & 69.11 & 68.38 & 82.67 &44.28\\
&zhihai & 37.50 & 20.98 & 4.59 & 13.69 & 64.54 & 67.75 & 77.68 & 76.99 & 77.16 &48.99\\
&LawGPT\_zh & 51.60 & 23.33 & 5.28 & 16.17 & 65.14 & 63.72 & 79.43 & 77.52 & \textbf{86.18} &52.04\\
&Hanfei & 51.12 & 23.95 & 5.19 & 18.76 & 65.12 & 70.83 & 75.01 & 74.21 & 76.97 &51.24\\
          \bottomrule
        \end{tabular}
    }
        \caption{Performance comparisons on the Layperson dataset. The best performance is indicated in bold. 
        }
    \label{tab:common results}
\end{table*}

We conduct extensive experiments on our \texttt{CitaLaw} using the proposed two-level evaluation methods.
\subsection{Experimental Settings}

\subsubsection{Evaluated Models}
We selected two categories of LLMs for testing: 
The legal LLMs include (1) \textbf{fuzi.mingcha} (6B)~\cite{sdu_fuzi_mingcha}, (2) \textbf{LexiLaw}\footnote{https://github.com/CSHaitao/LexiLaw} (6B), (3) \textbf{Tailing}\footnote{https://github.com/DUTIR-LegalIntelligence/Tailing} (7B), (4) \textbf{DISC-LawLLM} (13B)~\cite{yue2023disclawllm}, (5) \textbf{zhihai} (7B)~\cite{wisdomInterrogatory}, (6) \textbf{LawGPT\_zh} (6B)~\cite{LAWGPT-zh}, (7) \textbf{HanFei} (7B)~\cite{HanFei}. 
The open-domain LLMs include Qwen2 (7B)~\cite{qwen2} and Llama3 (8B)~\cite{llama3modelcard}. For these models, we tested all methods mentioned in Sec.~\ref{sec:method}, including:
(1) \textbf{CloseBook},  (2) \textbf{CGG}, (3) \textbf{ARG-Q} and (4) \textbf{ARG-QA}. 
For the legal LLMs, we generate responses using CGG. 
Appendix~\ref{sec:appendix evaluate models} has the details.

\subsubsection{Implementation Details}
Our implementation is based on the Huggingface Transformers library~\cite{wolf2020transformers} with PyTorch. We use bge-base-zh-v1.5~\cite{xiao2023c} as the retrieval model and conduct all experiments on Nvidia A6000 GPUs. For response generation, we set the temperature to 0 to ensure reproducibility. Additional details are provided in Appendix~\ref{sec:appendix details}.

\subsection{Main Results}

\begin{table*}[h!]
    \vspace{-3px}
    \resizebox{1\textwidth}{!}{
         \begin{tabular}{cl|c|ccc cccc|cccc|c}
          \toprule
          \multicolumn{2}{c|}{Metric} & \multicolumn{1}{c|}{Fluency } & \multicolumn{7}{c|}{Correctness} & \multicolumn{4}{c|}{Citation} &\multicolumn{1}{c}{All}\\ 
          \hline
          Category & Model & Mauve & Rouge-1 & Rouge-2 & Rouge-L & BERT-F & Correct$_\mathrm{c}$ & Correct$_\mathrm{a}$ & Correct$_\mathrm{d}$  & Cita$_\mathrm{Law}$ & Cita$_\mathrm{c}$ & Cita$_\mathrm{a}$ & Cita$_\mathrm{d}$ &\textbf{Avg}\\
          \hline
                    \multicolumn{1}{c}{\multirow {4}{*}{\shortstack{Llama3\\ (Llam3-8B-Instruct)}}}  
&CloseBook & 23.81 & 23.05 & 7.29 & 19.23 & 62.83 & \textbf{76.30} & 71.05 & 70.32 & 63.49 & 66.95 & 68.83 & 65.46 & 51.55 \\
&CGG & 36.37 & \textbf{26.15} & \textbf{7.84} & \textbf{19.55} & \textbf{65.60} & 67.19 & \textbf{76.36} & \textbf{77.73} & \textbf{73.58} & 68.23 & 67.87 & 67.65 & \textbf{54.51} \\
&ARG-Q & \textbf{42.65} & 20.39 & 5.07 & 15.75 & 62.82 & 70.49 & 73.67 & 72.00 & 68.61 & \textbf{69.48} & \textbf{70.51} & 68.34 & 53.31 \\
&ARG-QA & 36.94 & 18.64 & 4.56 & 14.63 & 61.50 & 71.07 & 72.38 & 70.32 & 69.40 & 68.95 & 70.42 & \textbf{69.51} & 52.36 \\
\hline
          \multicolumn{1}{c}{\multirow {4}{*}{\shortstack{Qwen2\\(Qwen2-7B-Instruct)}}}  
&CloseBook & \textbf{61.91} & 30.44 & 10.54 & \textbf{23.53} & 67.55 & \textbf{74.35} & 79.84 & 78.52 & 68.55 & 68.03 & 70.30 & 69.71 & \textbf{58.61} \\
&CGG & 39.66 & \textbf{31.01} & \textbf{10.75} & 23.43 & \textbf{69.06} & 73.49 & \textbf{80.11} & \textbf{81.11} & 70.37 & 67.82 & 69.53 & 70.01 & 57.20 \\
&ARG-Q & 41.02 & 20.57 & 5.14 & 15.62 & 63.31 & 67.84 & 74.71 & 73.94 & \textbf{73.01} & 68.96 & \textbf{73.20} & \textbf{73.64} & 54.25 \\
&ARG-QA & 21.97 & 16.67 & 3.06 & 12.47 & 60.70 & 67.49 & 71.16 & 70.88 & 71.76 & \textbf{69.01} & 71.04 & 71.33 & 50.63 \\
          \hline
          \multicolumn{1}{c}{\multirow {7}{*}{\shortstack{Legal LLM\\(CGG )}}}  
&DISC-LawLLM & 38.11 & 21.37 & 6.75 & 16.96 & 60.84 & \textbf{73.42} & 72.14 & 71.79 & 63.92 & 67.42 & 68.22 & 65.45 & 52.20 \\
&fuzi.mingcha & 66.55 & 28.95 & 9.51 & 22.69 & 67.06 & 70.73 & 76.66 & 77.47 & 65.92 & 66.94 & \textbf{69.28} & 68.69 & 57.54 \\
&LexiLaw & 57.74 & 29.01 & 8.93 & 23.83 & 65.63 & 70.36 & 76.67 & 75.97 & 65.28 & 66.93 & 68.89 & 68.03 & 56.44 \\
&Tailing & 50.16 & 26.52 & 9.16 & 22.44 & 65.35 & 75.96 & 73.83 & 70.30 & 64.65 & 66.94 & 67.56 & 66.09 & 54.91 \\
&zhihai & 26.29 & 21.38 & 6.00 & 15.53 & 64.47 & 65.59 & 76.38 & 77.37 & 67.93 & 66.30 & 63.17 & 59.82 & 50.85 \\
&LawGPT\_zh & 47.10 & 29.16 & 8.92 & 22.55 & 67.64 & 69.48 & \textbf{79.37} & \textbf{80.23} & 66.90 & \textbf{68.38} & 67.55 & \textbf{68.94} & 56.35 \\
&HanFei & \textbf{75.72} & \textbf{32.98} & \textbf{12.46} & \textbf{26.91} & \textbf{68.72} & 73.25 & 78.63 & 78.11 & \textbf{67.03} & 67.45 & 68.63 & 67.73 & \textbf{59.80} \\
          \bottomrule
        \end{tabular}
    }
        \caption{Performance comparisons on the Practitioner dataset. The best performance is indicated in bold. 
        }
    \vspace{-3mm}
    \label{tab:legal results}
\end{table*}

The results on the Layperson and Practitioner datasets are presented in Table~\ref{tab:common results} and Table~\ref{tab:legal results}.
We analyze the results from three perspectives:

\subsubsection{Peformance of Open-Domain LLM}

\textbf{Legal references improve the response quality.}
Compared to CloseBook, the overall performance in CGG, ARG-Q, and ARG-QA has improved. This indicates that incorporating references into the LLM helps it better understand both the question and the required direction for the answer, thereby enhancing performance in terms of style consistency, correctness, and citation quality.

\textbf{CGG achieves better response quality.}
We observe that CGG achieves optimal performance, especially response correctness, suggesting that incorporating legal references into the LLM input is more effective than refining the LLM’s response. This is because including legal knowledge as input allows the LLM to consider relevant context when generating replies, whereas refining the response might lead to excessive alterations.

\textbf{ARG improves the alignment of responses and references.}
We can observe that ARG outperforms CGG in citation-related metrics overall. This is because CGG merely incorporates reference information as input, which may lead the model to overlook some reference details during the generation process. In contrast, ARG modifies the answer based on the references after generation, making it easier to ensure the completeness of citations.

\textbf{Chinese data fine-tuning can bring benefits.}
Both the Layperson and Practitioner datasets are Chinese datasets. Qwen2 (Fine-tuning on more Chinese data) achieved better performance than Llama3, demonstrating the benefits of using Chinese data for fine-tuning.

\textbf{CloseBook tends to state circumstances.}
CloseBook performs better in terms of correctness regarding circumstances compared to the other dimensions. This suggests that when judicial knowledge references are not used, the LLM is more likely to repeat the circumstances itself, rather than providing an appropriate response to the illegal acts and the legal decision.

\subsubsection{Performance of Legal LLM}
\textbf{Law article training achieves gains.}
In the Layperson dataset, LexiLaw achieves optimal performance overall. This is because the questions in the Layperson dataset often require only law articles to provide answers clearly, and LexiLaw’s training explicitly used law articles, allowing it to effectively handle such questions.

\textbf{Full-parameter training offers advantages.}
Hanfei achieves the best results in the Practitioner dataset, as it is a fully parameter-trained legal LLM. Full-parameter fine-tuning allows it to effectively simulate a legal expert, thus performing well. 

\textbf{Syllogistic reasoning is useful.}
fuzi.mingcha performs well on syllogism evaluation metrics, particularly on the Layperson dataset. This is due to its fine-tuning of syllogism judgment data. 

\subsubsection{Open Domain LLM vs. Legal LLM}
\textbf{Impact of LLM Backbone.}
We can observe that some legal LLMs perform worse than open-domain LLMs. This is because Qwen2 and Llama3 are the latest open-domain LLMs, and their overall capabilities have significantly improved. In contrast, most legal LLMs are built on earlier generations of LLMs, which have weaker base models, leading to poorer overall performance.

\textbf{Effectiveness of legal knowledge.}
Overall, 
the upper limit of legal LLMs is higher than that of open-domain LLMs. 
This is because legal LLMs, after extensive training on legal knowledge, have developed strong capabilities in solving legal issues. As a result, even though their base models are outdated, they can still perform effectively.

\subsection{Human Evaluation}\label{sec:human}
In this section, we compared the syllogism-level metric with human evaluation. Details of legal human annotators can be found in Appendix~\ref{sec:appendix entailment}.
The syllogism-level evaluation of citation quality is divided into two stages:
Stage 1: Extracting key components. Stage 2: Assessing the entailment using an NLI model.

\textbf{Stage 1:} We randomly selected 50 questions each from the Layperson and Practitioner datasets. 
After splitting the cases into individual clauses, annotators were provided with the full case and its clauses. 
They do a three-class classification of each clause.
The Qwen2's annotations were then compared with human annotations. 
The Cohen's kappa coefficient~\cite{cohen1960coefficient} of 0.7876 indicates substantial agreement (0.61–0.80) between the model's and human annotators' labels.

\textbf{Stage 2:} We randomly selected 50 questions from the Practitioner dataset and used Qwen2 to extract key components of pairs of responses and citations. 
Annotators assessed the degree to which the citations entailed the corresponding response components using a 5-point scale (1: low, 5: high), with descriptions provided in Appendix~\ref{sec:appendix entailment}.
The entailment probabilities given by DISC-LawLLM, which range from 0 to 1, were scaled to the same 1–5 range by multiplying by 5 and rounding. We then compared the scaled model outputs with the human evaluations and calculated Cohen’s kappa coefficient.
The kappa score of 0.6923 again indicates substantial agreement (0.61–0.80) between the model and human judgments.

\begin{figure}
    \centering
\includegraphics[width=0.98\linewidth]{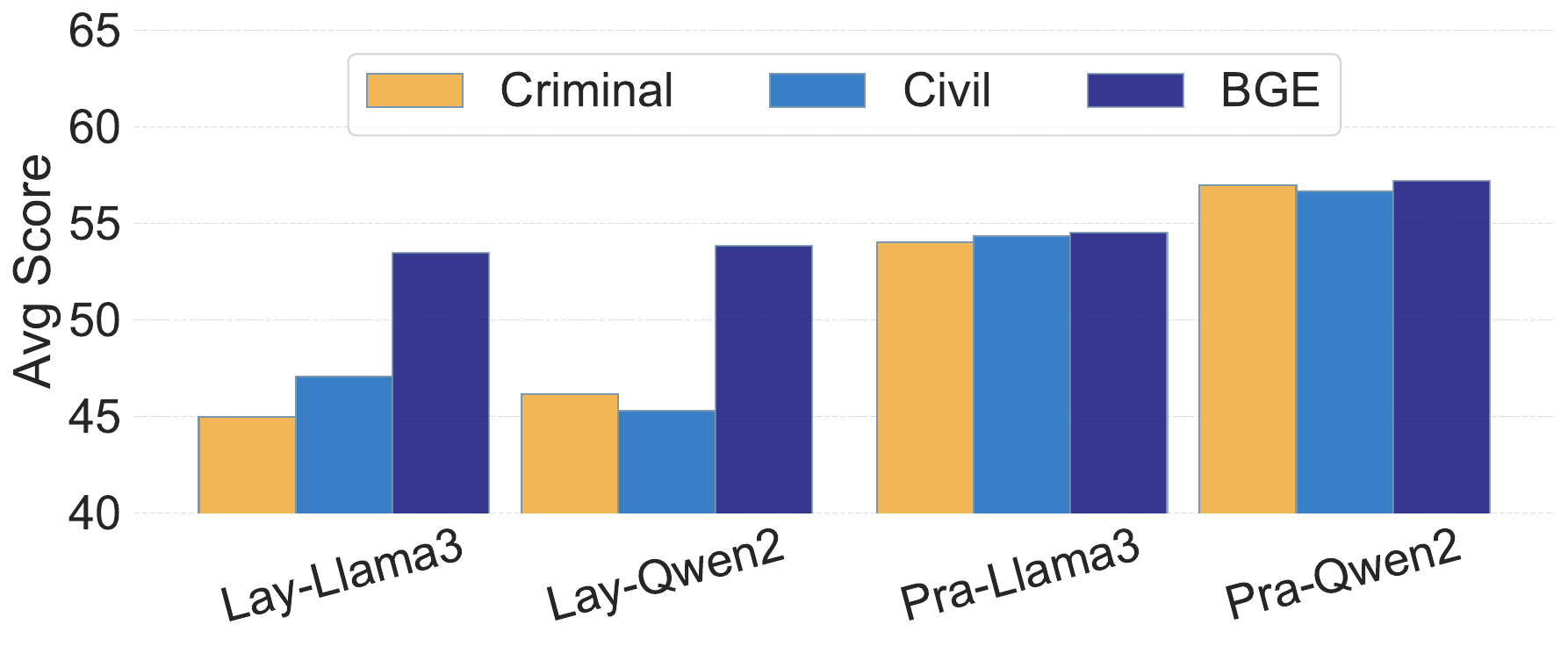}
    \caption{
Performance of different retrieval models. Lay is short for Layperson dataset and Pra is short for Practitioner dataset.
            }
    \label{fig:retrieve model}
    \vspace{-3mm}
\end{figure}

\subsection{Effects on Different Retrieval Models}\label{sec:different retrieval model}
We selected BGE as the retrieval model in the main experiment.  
In this section, we explore the impact of using different retrieval models. Specifically, we evaluate Criminal-BERT~\cite{zhong2019openclap} and Civil-BERT~\cite{zhong2019openclap}, two legal domain models based on BERT, fine-tuned on large-scale criminal and civil law documents, respectively. 
We replaced the retrieval model and tested the CGG method on the Layperson dataset. The average results across all metrics are shown in Figure~\ref{fig:retrieve model}, with detailed metric results provided in Appendix~\ref{sec:appendix retrieval model}.

As shown, on the Layperson dataset, BGE significantly outperforms the other two models. This is because the dataset consists of questions from laypersons, which are more everyday in nature. In contrast, the two legal BERT models, having been trained extensively on legal cases, show a distributional mismatch with open-domain data, leading to poorer performance.
On the Practitioner dataset, which features professional legal questions, BGE still achieves the best performance. This can be attributed to its extensive training on diverse data, likely including some legal data, and its use of more advanced model architectures and techniques. However, the two legal BERT models perform comparably to BGE, showcasing the benefits of their specialized training on legal data.

\subsection{Effects on Different NLI Models}\label{sec:different nli}

\begin{figure}[t]
    \centering

            \begin{subfigure}{0.98\linewidth}
        \centering
    \includegraphics[width=\textwidth]{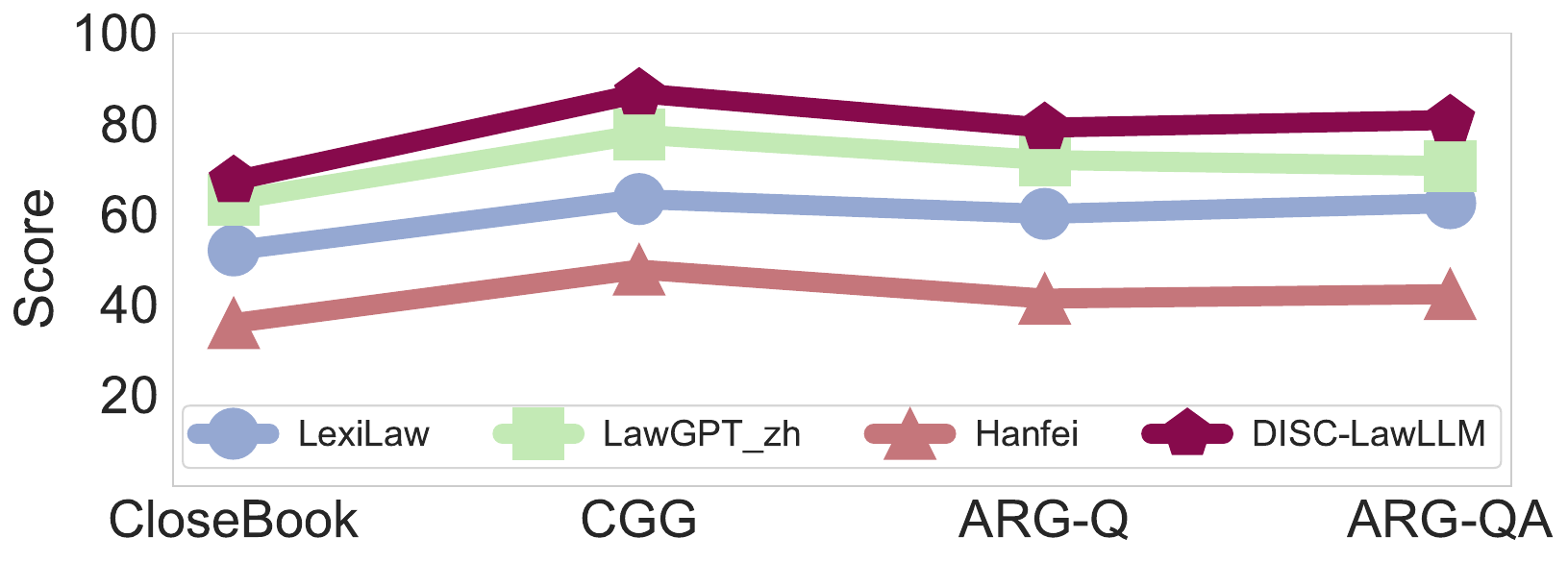}
        \subcaption{
        Methods for Cita$_\mathrm{Law}$ metric with Layperson dataset.
        }
    \end{subfigure}
    
        \begin{subfigure}{0.98\linewidth}
        \centering
    \includegraphics[width=\textwidth]{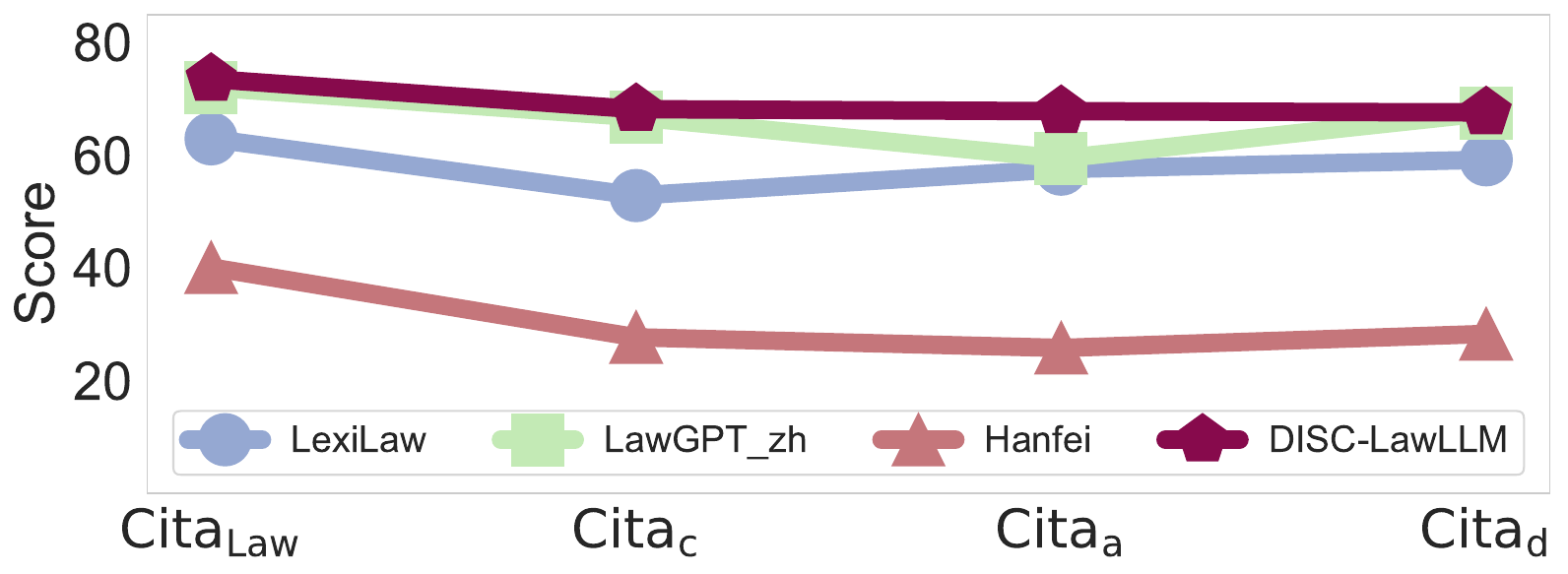}
        \subcaption{
        Metrics for CGG method with Practitioner dataset.
        }
    \end{subfigure}
    
    \caption{
    The performance of different NLI models when the LLM is Llama.
    }
\label{fig:nli llama}
\vspace{-4mm}
\end{figure}

We opted to use legal LLMs as the NLI model in our experiments, as they support longer input lengths and incorporate substantial legal knowledge. 
In Section~\ref{sec:human}, we verified that DISC-LawLLM and human achieved good consistency.
In this section, we explore the performance of several legal LLMs in the NLI task. Besides DISC-LawLLM, we evaluated LexiLaw, LawGPT\_zh, and Hanfei, which demonstrated strong performance in the main experiments. 

In Figures~\ref{fig:nli llama} (a), we examined the ability of four legal LLMs to evaluate Llama across the CloseBook, CGG, ARG-Q, and ARG-QA methods using the Cita$_\mathrm{Law}$ metric on the Layperson dataset. 
In Figures~\ref{fig:nli llama} (b), we investigated the performance of four legal LLMs in evaluating the CGG method applied to Llama across the metrics Cita$_\mathrm{Law}$, Cita$_\mathrm{c}$, Cita$_\mathrm{a}$, and Cita$_\mathrm{d}$ on the Practitioner dataset.

We can observe that Hanfei provides lower entailment scores across both datasets. This is because it is a fully parameter-tuned legal LLM, which results in a diminished capability to handle the general task of entailment reasoning.
Additionally, we found that on the Practitioner dataset, other legal LLMs achieved results closer to those of DISC-LawLLM, while on the Layperson dataset, the performance gap was significantly larger.
This is because the Practitioner dataset is more judicially oriented, aligning with the knowledge seen during the fine-tuning of legal LLMs. In contrast, due to limited training on general-purpose data, other legal LLMs struggle to accurately determine entailment relationships in the Layperson dataset.
Similar conclusions can be drawn when the LLM is Qwen in Appendix~\ref{sec:appendix nli model}.

\section{Conclution}
We introduce \texttt{CitaLaw}, a benchmark designed to explore LLMs to generate responses with citations in legal scenarios, thus improving the trustworthiness of LLMs.
\texttt{CitaLaw} includes two categories of questions: laypersons and practitioners. For laypersons, \texttt{CitaLaw} provides law articles as citations to help them understand the LLM’s response clearly. For practitioners, both law articles and precedent cases are provided as citations, better supporting their needs for complex reasoning.  
\texttt{CitaLaw} offers global-level and syllogism-level metrics and supports the integration of citations into LLM inputs to guide generation or using citations to refine LLM's response.
We conducted extensive experiments on 7 legal-domain LLMs and 2 popular open-domain LLMs, 
providing valuable insights for the deployment of LLMs in legal scenarios.

\section{Limitations}
While \texttt{CitaLaw} provides a robust framework for evaluating LLMs in legal scenarios, several limitations should be acknowledged to guide future extensions of this work.

First, the datasets used in \texttt{CitaLaw} are primarily sourced from the Chinese legal system, which may limit the benchmark’s applicability to other jurisdictions. However, by incorporating both law articles and precedent cases to align with the principles of civil and common law systems, \texttt{CitaLaw} demonstrates strong potential for adaptation to diverse legal contexts.

Second, the syllogism-based evaluation framework simplifies legal reasoning into three key components: the major premise (law articles or precedent cases), the minor premise (case circumstances and actions), and the conclusion (legal decision). While this structured approach is effective for systematic evaluation, real-world legal reasoning may encompass additional complexities. 

\section{Ethical Considerations}
\textbf{Data Privacy and Confidentiality.}
The legal datasets used in \texttt{CitaLaw} include law articles, precedent cases, user questions, and golden responses. These documents were sourced from publicly available databases, ensuring compliance with data privacy and confidentiality standards. We carefully reviewed the datasets to ensure that no personally identifiable information (PII) or sensitive details about individuals were inadvertently included. 

\textbf{Alignment with Legal Standards.}
Legal AI systems must align with the ethical and professional standards of the legal domain. Our work emphasizes the need for syllogism-based reasoning to ensure logical consistency and adherence to legal principles.

\textbf{Transparency and Explainability.}
Legal reasoning must be transparent and interpretable, particularly when used in sensitive or high-stakes domains. The metrics proposed in \texttt{CitaLaw}, including syllogism-based evaluation, aim to improve explainability by breaking down the reasoning process into logical components. 

\textbf{Responsibility in System Deployment.}
\texttt{CitaLaw} is intended as a research benchmark and should not be directly deployed in high-stakes legal decision-making without human oversight. While the benchmark aims to enhance the trustworthiness of LLM-generated responses, legal professionals should always verify the citations and legal interpretations provided by such systems. Misuse of automated systems without adequate validation could lead to inaccurate legal advice or unintended consequences in legal proceedings.

\bibliography{custom}

\clearpage
\appendix

\section{The Used Prompts}\label{sec:appendix:prompt}
\begin{figure*}
    \centering
\includegraphics[width=0.98\linewidth]{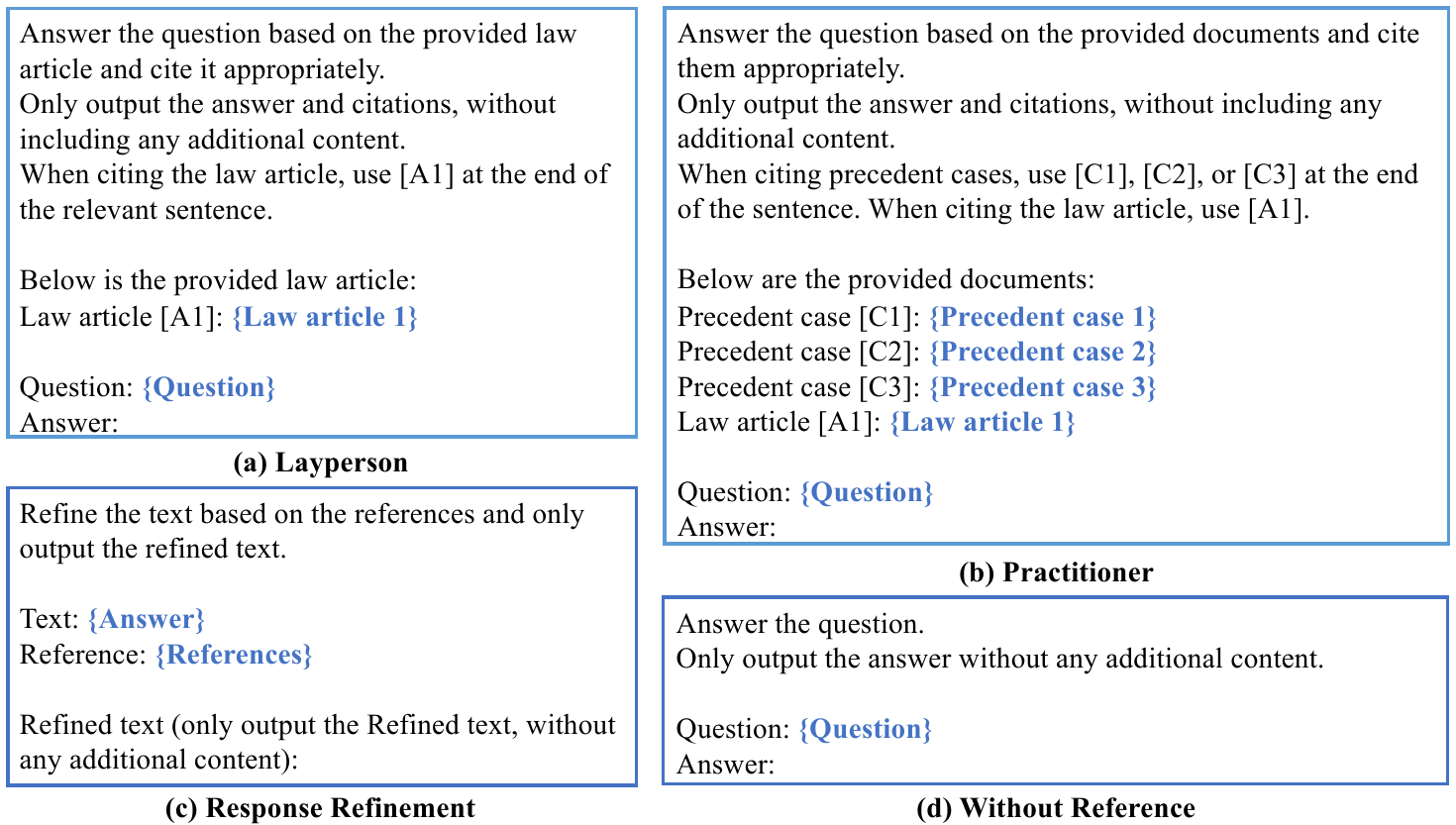}
    \caption{
    Prompts used in this paper. (a) The prompt $p_1$ is used to retrieve one law article in the Layperson dataset.
(b) The prompt $p_1$ is used to retrieve one law article and three precedent cases in the Practitioner dataset.
(c) The prompt $p_3$ is used to refine the LLM's answer based on references.
(d) The prompt $p_2$ is used for LLM responses without references.
            }
    \label{fig:prompt}
    \vspace{-3mm}
\end{figure*}
Figure~\ref{fig:prompt} illustrates the prompts used in this paper, including $p_1$, $p_2$, $p_3$ in Eq.~\ref{eq:type1}, Eq.~\ref{eq:type2} and Eq.~\ref{eq:type3}.

\section{More Details of Evaluated Models and Datasets}\label{sec:appendix evaluate models}
For the Legal LLMs, we choose
(1) \textbf{fuzi.mingcha} (6B)~\cite{sdu_fuzi_mingcha}: 
It leverages unsupervised judicial corpora for training and uses syllogistic reasoning judgment data for fine-tuning. (2) \textbf{LexiLaw}\footnote{https://github.com/CSHaitao/LexiLaw} (6B): It specifically utilizes legal articles and legal reference books for training. (3) \textbf{Tailing}\footnote{https://github.com/DUTIR-LegalIntelligence/Tailing} (7B): It uses judicial text validation data, information extraction data, and judgment data for training. (4) \textbf{DISC-LawLLM} (13B)~\cite{yue2023disclawllm}: In addition to fine-tuning with pairs, it also uses triplet data for fine-tuning to enhance the model's ability to leverage external knowledge. (5) \textbf{zhihai} (7B)~\cite{wisdomInterrogatory}: It utilizes ChatGPT to modify the existing dataset and then performs secondary pre-training. (6) \textbf{LawGPT\_zh} (6B)~\cite{LAWGPT-zh}: It primarily uses scenario-based dialogues and knowledge-based question-answering data for fine-tuning based on LoRA.  (7) \textbf{HanFei} (7B)~\cite{HanFei}: It is the first fully parameter-trained legal LLM in China. 
Because in the main experiment, CGG has the best overall performance, for the legal LLMs, we generate responses using CGG.

Table~\ref{tab:model urls and license} and Table~\ref{tab:dataset urls and license} are the website URLs and corresponding licenses of the evaluated models and datasets.

\begin{table*}[t]
\centering
\resizebox{0.9\textwidth}{!}{
\begin{tabular}{cl|ll}
    \toprule
    Type&LLM &URL & Licence   \\
    \hline
    \multicolumn{1}{c}{\multirow {2}{*}{\shortstack{Open domain}}}&
Qwen2-7B-Instruct &\url{https://huggingface.co/Qwen/Qwen2-7B-Instruct}&Apache-2.0 license\\ 
    &Llam3-8B-Instruct &\url{https://github.com/meta-llama/llama3} &META LLAMA 3 COMMUNITY License\\
    \hline
    \multicolumn{1}{c}{\multirow {7}{*}{\shortstack{Legal Domain}}}
&fuzi.mingcha&\url{https://github.com/irlab-sdu/fuzi.mingcha} &Apache-2.0 license\\
    &DISC-LawLLM &\url{https://github.com/FudanDISC/DISC-LawLLM}&Apache-2.0 license \\
    &LawGPT\_zh&\url{https://github.com/LiuHC0428/LAW-GPT}& \\
    &Hanfei&\url{https://github.com/siat-nlp/HanFei}&Apache-2.0 license\\
    &Tailing&\url{https://github.com/DUTIR-LegalIntelligence/Tailing}&\\
    &LexiLaw&\url{https://github.com/CSHaitao/LexiLaw}&MIT license\\
    &zhihai&\url{https://github.com/zhihaiLLM/wisdomInterrogatory}&Apache-2.0 license\\
    
    \bottomrule
\end{tabular}
}
\caption{
The LLM source URLs and licenses used by CitaLaw. The parts where the license is listed as empty indicate that the author has not provided a License.}
\label{tab:model urls and license}
\end{table*}

\begin{table*}[t]
\centering
\resizebox{0.8\textwidth}{!}{
\begin{tabular}{cl|ll}
    \toprule
    Type&Dataset &URL & Licence   \\
    \hline
    \multicolumn{1}{c}{\multirow {2}{*}{\shortstack{Question}}}&
    Layperson &\url{https://github.com/open-compass/LawBench}&Apache-2.0 license\\ 
    &Practitioner &\url{https://github.com/CSHaitao/LexEval} &MIT License\\
    \hline
    \multicolumn{1}{c}{\multirow {8}{*}{\shortstack{Corpus}}}&
LeCaRD&\url{https://github.com/myx666/LeCaRD}&MIT License\\
    &ELAM&\url{https://github.com/ruc-wjyu/IOT-Match}&MIT License \\
    &CAIL2021-sfzy&\url{https://github.com/china-ai-law-challenge/CAIL2021}&\\
    &LJP-MSJudg&\url{https://github.com/mly-nlp/LJP-MSJudge}&\\
    &fuzi.mingcha&\url{https://github.com/irlab-sdu/fuzi.mingcha} &Apache-2.0 license\\
    &DISC-LawLLM &\url{https://github.com/FudanDISC/DISC-LawLLM}&Apache-2.0 license \\
    &LawGPT\_zh&\url{https://github.com/LiuHC0428/LAW-GPT}& \\
    &Hanfei&\url{https://github.com/siat-nlp/HanFei}&Apache-2.0 license\\
    \bottomrule
\end{tabular}
}
\caption{
The dataset source URLs and licenses used by CitaLaw. The parts where the license is listed as empty indicate that the author has not provided a License.}
\label{tab:dataset urls and license}
\end{table*}

\section{More Details on Implementation}\label{sec:appendix details}
Considering the length of legal texts and the input window for the LLMs is limited, all experiments in this paper are conducted using a zero-shot setting.
We use the Chinese-performing-well Qwen2-1.5B~\cite{qwen2}\footnote{https://huggingface.co/Qwen/Qwen2-1.5B} to complete the MAUVE calculations. For RGUGE, We use version 1.0.1 of ROUGE for calculation. For BERTScore, we use bert-base-chinese~\cite{devlin2018bert}\footnote{https://huggingface.co/google-bert/bert-base-chinese} to compute it.
Regarding sentence-BERT, we employ paraphrase-multilingual-MiniLM-L12-v2~\cite{reimers2019sentence}\footnote{https://huggingface.co/sentence-transformers/paraphrase-multilingual-MiniLM-L12-v2}.

\section{Human Evaluation}\label{sec:appendix entailment}
We hired four legal annotators from a Chinese university, all of whom have legal education backgrounds and are familiar with the cases in the dataset they need to annotate. 
We explained to the annotators that the data they annotated would be used for scientific research and paid them a reasonable remuneration based on local conditions.
They are all graduate students from the judicial field, with practical experience in the legal profession. Two are male, two are female, aged between 24 and 30, and all have over five years of judicial theory study. Two annotators were responsible for the first stage of annotation, while the other two were responsible for the second stage, with all working together on the annotation process.

Table~\ref{tab:entail description} shows a detailed description of each level used to evaluate the agreement of the NLI model with human evaluations.

\begin{table*}[t]
\centering
\resizebox{1\textwidth}{!}{
\begin{tabular}{ll}
    \toprule
    \multirow{1}{*}{\textbf{Score}} &\multirow{1}{*}{\textbf{Description}} \\
    \midrule
    \textbf{1 } & No Entailment: The former does not entail the latter at all, with no logical connection between the two. \\
    \textbf{2 } &Weak Entailment: A partial entailment where the former somewhat relates to the latter, but the connection is weak and not fully conclusive.\\
    \textbf{3 }&Moderate Entailment: A moderate degree of entailment, meaning the former generally leads to the latter in most cases, but exceptions exist. \\
    \textbf{4 }& Strong Entailment: A strong logical relationship where the former can derive the latter in the vast majority of cases.\\
    \textbf{5 }& Complete Entailment: The former fully entails the latter in all cases, with an unambiguous and definitive logical connection between them.\\
    \bottomrule
\end{tabular}}
\caption{Scoring Criteria for Human Evaluation of Entailment.}
\label{tab:entail description}
\end{table*}

\section{Different Retrieval Models}\label{sec:appendix retrieval model}

\begin{table*}[h!]
    \vspace{-3px}
    \resizebox{1\textwidth}{!}{
         \begin{tabular}{cl|c|ccc cccc|c|c}
          \toprule
          \multicolumn{2}{c|}{Metric} & \multicolumn{1}{c|}{Fluency } & \multicolumn{7}{c|}{Correctness} & \multicolumn{1}{c|}{Citation}  & \multicolumn{1}{c}{All}\\
          \hline
          Category & Retriever  & Mauve & Rouge-1 & Rouge-2 & Rouge-L & BERT-F & Correct$_c$ & Correct$_a$ & Correct$_d$  &  Cita$_{Law}$  &\textbf{Avg }\\
          \hline
\multicolumn{1}{c}{\multirow {3}{*}{\shortstack{Llama3\\ (Llam3-8B-Instruct)}}}  
            &Criminal & 37.44 & 18.07 & 2.18 & 13.15 & 61.71 & 64.03 & 63.56 & 64.36 & 80.34 & 44.98 \\
&Civil & 56.16 & 18.27 & 2.34 & 13.44 & 61.90 & 63.22 & 63.89 & 63.35 & 80.97 & 47.06\\
&BGE & 61.01 & 23.97 & 6.05 & 17.91 & 65.94 & 67.29 & 77.31 & 74.95 & 86.70 & \textbf{53.46} \\
\hline
\multicolumn{1}{c}{\multirow {3}{*}{\shortstack{Qwen2\\(Qwen2-7B-Instruct)}}}  
            &Criminal & 55.26 & 21.09 & 4.53 & 14.32 & 64.73 & 63.10 & 64.89 & 65.85 & 61.60 & 46.15 \\
&Civil & 52.44 & 20.48 & 4.16 & 13.81 & 64.45 & 61.79 & 64.94 & 65.62 & 59.88 & 45.29\\
&BGE & 75.10 & 22.26 & 4.77 & 15.41 & 65.28 & 67.50 & 78.62 & 77.82 & 77.59 & \textbf{53.82} \\
          \bottomrule
        \end{tabular}
    }
        \caption{Performance comparisons on retrieval models in the Layperson dataset when the method is CGG. The best performance is indicated in bold. 
        }
    \vspace{-3mm}
    \label{tab:retrieve model common results}
\end{table*}

\begin{table*}[h!]
    \vspace{-3px}
    \resizebox{1\textwidth}{!}{
         \begin{tabular}{cl|c|ccc cccc|cccc|c}
          \toprule
          \multicolumn{2}{c|}{Metric} & \multicolumn{1}{c|}{Fluency } & \multicolumn{7}{c|}{Correctness} & \multicolumn{4}{c|}{Citation}  & \multicolumn{1}{c}{All}\\
          \hline
          Category & Retriever  & Mauve & Rouge-1 & Rouge-2 & Rouge-L & BERT-F & Correct$_c$ & Correct$_a$ & Correct$_d$  &  Cita$_{Law}$ & Cita$_c$ & Cita$_a$ & Cita$_d$ &\textbf{Avg }\\
          \hline
\multicolumn{1}{c}{\multirow {3}{*}{\shortstack{Llama3\\ (Llam3-8B-Instruct)}}}  
&Criminal & 34.25 & 25.79 & 7.86 & 19.42 & 65.03 & 66.27 & 76.30 & 76.82 & 70.59 & 66.41 & 70.09 & 69.47 & 54.03 \\
&Civil & 39.84 & 26.39 & 8.07 & 20.02 & 65.27 & 65.41 & 75.78 & 75.73 & 69.21 & 67.52 & 69.54 & 69.16 & 54.33 \\
&BGE & 36.37 & 26.15 & 7.84 & 19.55 & 65.60 & 67.19 & 76.36 & 77.73 & 73.58 & 68.23 & 67.87 & 67.65 & \textbf{54.51} \\
\hline
\multicolumn{1}{c}{\multirow {3}{*}{\shortstack{Qwen2\\(Qwen2-7B-Instruct)}}}  
&Criminal & 32.49 & 31.79 & 11.09 & 23.93 & 69.79 & 72.00 & 80.81 & 81.53 & 68.42 & 68.42 & 71.86 & 71.54 & 56.97 \\
&Civil & 33.37 & 31.67 & 11.06 & 23.84 & 69.63 & 73.35 & 80.57 & 81.27 & 69.11 & 66.41 & 70.09 & 69.47 & 56.65 \\
&BGE & 39.66 & 31.01 & 10.75 & 23.43 & 69.06 & 73.49 & 80.11 & 81.11 & 70.37 & 67.82 & 69.53 & 70.01 & \textbf{57.20} \\
          \bottomrule
        \end{tabular}
    }
        \caption{Performance comparisons on retrieval models in the Practitioner dataset when the method is CGG. The best performance is indicated in bold. 
        }
    \vspace{-3mm}
    \label{tab:retrieve model legal results}
\end{table*}

Tables~\ref{tab:retrieve model common results} and \ref{tab:retrieve model legal results} present the performance of different retrieval models—Criminal-BERT, Civil-BERT, and BGE—on each metric for the CGG method across the two datasets. 
It can be observed that when Llama3 and Qwen2 are used as LLMs, BGE achieves the best performance as the retrieval model. Comparing the two datasets, on the Layperson dataset, where the questions are more general, Criminal-BERT and Civil-BERT, which focus on legal cases, perform relatively poorly. In contrast, on the Practitioner dataset, despite no structural or training improvements, Criminal-BERT and Civil-BERT achieve results comparable to BGE, highlighting the importance of legal knowledge in judicial QA tasks.

The differences between the two datasets also underscore the significance of selecting an appropriate retrieval model.

\section{Different NLI Models}\label{sec:appendix nli model}
\begin{figure}[t]
    \centering
        \begin{subfigure}{0.98\linewidth}
        \centering
    \includegraphics[width=\textwidth]{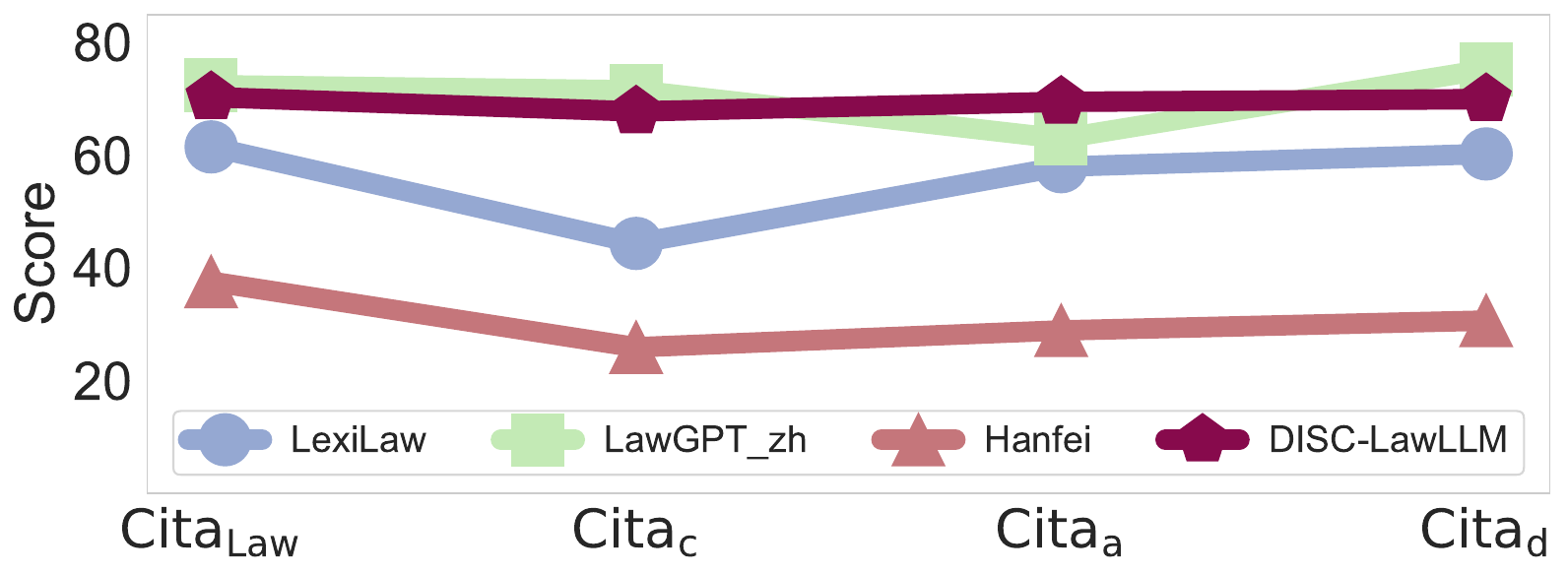}
        \subcaption{Metrics for CGG  method with Layperson dataset.}
    \end{subfigure}
    
            \begin{subfigure}{0.98\linewidth}
        \centering
    \includegraphics[width=\textwidth]{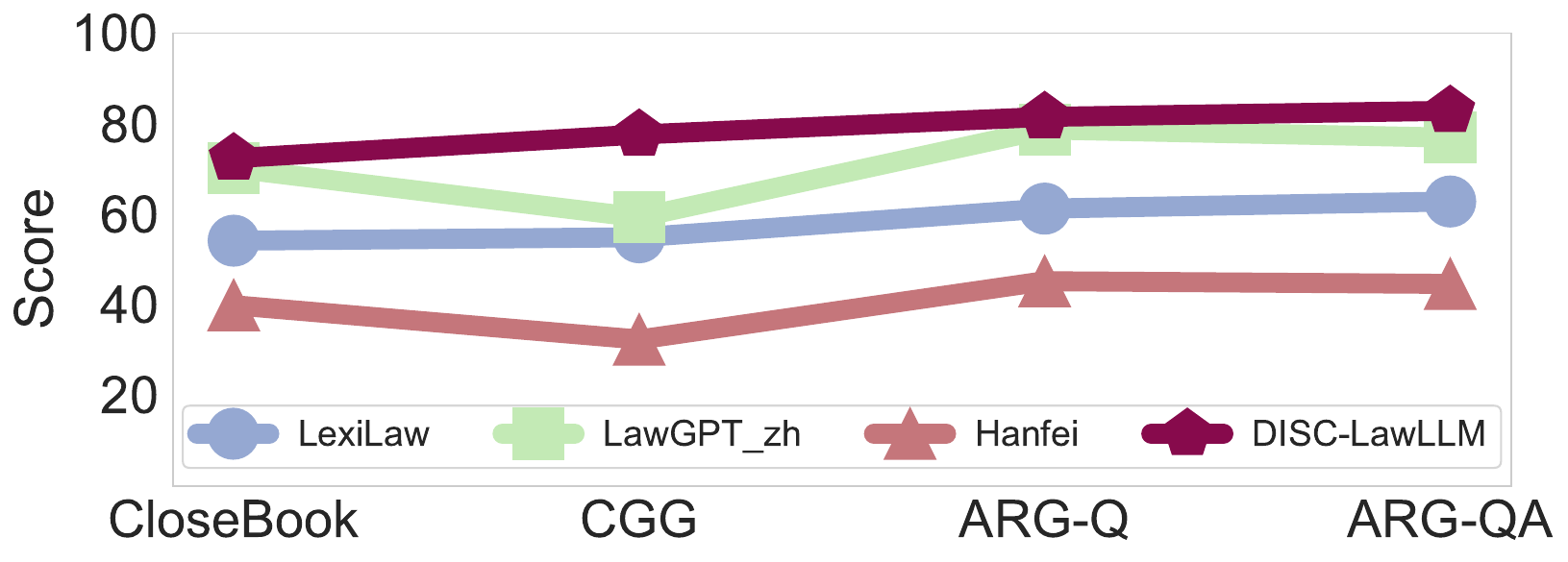}
        \subcaption{Methods for Cita$_\mathrm{Law}$ metric with Practitioner dataset.}
    \end{subfigure}
    
    \caption{
    The performance of different NLI models when the LLM is Qwen.
}

\label{fig:nli qwen}
\end{figure}

Figures~\ref{fig:nli qwen} (a) and (b) show the entailment scores given by four legal LLMs as NLI models under different methods (CloseBook, CGG, ARG-Q, ARG-QA) and metrics(Cita$_\mathrm{Law}$, Cita$_\mathrm{S}$, Cita$_\mathrm{B}$, and Cita$_\mathrm{C}$) when Qwen is used as the LLM. 
Similar conclusions to those in Section~\ref{sec:different nli} can be drawn.




\end{document}